\documentclass[conference]{IEEEtran}
\IEEEoverridecommandlockouts
\usepackage{amsmath,amssymb,amsfonts}
\usepackage{algorithmic}
\usepackage{graphicx}
\usepackage{textcomp}
\usepackage{xcolor}

\usepackage{adjustbox}
\usepackage{subfig}
\usepackage{makecell}
\usepackage{multirow}
\usepackage{booktabs}
\usepackage{newfloat}
\usepackage[hyphens]{url}
\usepackage{algorithm}
\usepackage{listings}

\usepackage{natbib}
\setcitestyle{numbers,square} %Citation-related commands

\def\BibTeX{{\rm B\kern-.05em{\sc i\kern-.025em b}\kern-.08em
    T\kern-.1667em\lower.7ex\hbox{E}\kern-.125emX}}

\begin{document}

\title{A $K$-variate Time Series Is Worth $K$ Words:\\ Evolution of the Vanilla Transformer Architecture for Long-term Multivariate Time Series Forecasting
}

\author{
    \IEEEauthorblockN{Zanwei Zhou$^1$, Ruizhe Zhong$^{1}$, Chen Yang$^1$, Yan Wang$^2$, Xiaokang Yang$^{1}$, Wei Shen$^{1*}$\thanks{* Corresponding Author} }
    \IEEEauthorblockA{$^1$MoE Key Lab of Artificial Intelligence, AI Institute, Shanghai Jiao Tong University, China \\
    $^2$School of Communication and Electronic Engineering, China\\
    \{SJTU19zzw, zerzerzerz271828, ycyangchen, xkyang, wei.shen\}@sjtu.edu.cn, ywang@cee.ecnu.edu.cn
    }
}

% \author{\IEEEauthorblockN{1\textsuperscript{st} Zanwei Zhou}
% \IEEEauthorblockA{\textit{MoE Key Lab of Artificial Intelligence, AI Institute} \\
% \textit{Shanghai Jiao Tong University}\\
% Shanghai, China \\
% SJTU19zzw@sjtu.edu.cn}
% \and
% \IEEEauthorblockN{2\textsuperscript{nd} Ruizhe Zhong}
% \IEEEauthorblockA{\textit{MoE Key Lab of Artificial Intelligence, AI Institute} \\
% \textit{Shanghai Jiao Tong University}\\
% Shanghai, China \\
% zerzerzerz271828@sjtu.edu.cn}
% \and
% \IEEEauthorblockN{3\textsuperscript{rd} Chen Yang}
% \IEEEauthorblockA{\textit{MoE Key Lab of Artificial Intelligence, AI Institute} \\
% \textit{Shanghai Jiao Tong University}\\
% Shanghai, China \\
% ycyangchen@sjtu.edu.cn}
% \and
% \IEEEauthorblockN{4\textsuperscript{th} Yan Wang}
% \IEEEauthorblockA{\textit{School of Communication and Electronic Engineering} \\
% \textit{East China Normal University}\\
% Shanghai, China \\
% ywang@cee.ecnu.edu.cn}
% \and
% \IEEEauthorblockN{5\textsuperscript{th} Xiaokang Yang}
% \IEEEauthorblockA{\textit{MoE Key Lab of Artificial Intelligence, AI Institute} \\
% \textit{Shanghai Jiao Tong University}\\
% Shanghai, China \\
% xkyang@sjtu.edu.cn}
% \and
% \IEEEauthorblockN{6\textsuperscript{th} Wei Shen}
% \IEEEauthorblockA{\textit{MoE Key Lab of Artificial Intelligence, AI Institute} \\
% \textit{Shanghai Jiao Tong University}\\
% Shanghai, China \\
% wei.shen@sjtu.edu.cn}
% }

\maketitle

\begin{abstract}
    % Multivariate time series forecasting (MTSF) is a fundamental problem in numerous real-world applications. Recently, Transformer has become the \emph{de facto} solution for MTSF, especially for the long-term cases. However, the basic configurations in existing MTSF Transformer architectures were barely carefully verified. In this study, we point out that the current tokenization strategy in MTSF Transformer architectures ignores the core characteristics of a time series which are its trend and seasonality rather than some scattered points. Therefore, the vanilla MTSF transformer struggles to capture the long-range dependency in time series and presents inferior performance. Based on this observation, we make a series of evolution on the basic configurations of the vanilla MTSF transformer. We  explore different tokenization strategies of a vanilla MTSF transformer, along with the encoder-decoder structure and embeddings. Surprisingly, the evolved transformer architecture is very simple yet highly effective, which is significantly superior to the vanilla Transformer and even substantially outperforms the state-of-the-art Transformers that are well-designed for MTSF. Furthermore, this architecture is extensible and can be enhanced by some time-series representations, e.g., the Fourier transform. %Code will be released soon.
    
    Multivariate time series forecasting (MTSF) is a fundamental problem in numerous real-world applications. Recently, Transformer has become the \emph{de facto} solution for MTSF, especially for the long-term cases. However, except for the one forward operation, the basic configurations in existing MTSF Transformer architectures were barely carefully verified. In this study, we point out that the current tokenization strategy in MTSF Transformer architectures ignores the \textit{token uniformity} inductive bias of Transformers. Therefore, the vanilla MTSF transformer struggles to capture details in time series and presents inferior performance. Based on this observation, we make a series of evolution on the basic architecture of the vanilla MTSF transformer. We vary the flawed tokenization strategy, along with the decoder structure and embeddings. Surprisingly, the evolved simple transformer architecture is highly effective, which successfully avoids the over-smoothing phenomena in the vanilla MTSF transformer, achieves a more detailed and accurate prediction, and even substantially outperforms the state-of-the-art Transformers that are well-designed for MTSF.  %Code will be released soon.
\end{abstract}

\section{Introduction}

A Multivariate time series consists of multiple intrinsically related time series variables. With the explosive growth of portable sensors and big data, the multivariate time series is becoming ubiquitous in many fields, e.g., power dispatch, financial transactions, traffic regulation, and weather forecasting. 
Therefore, the multivariate time-series forecasting (MTSF) problem has been gaining increasing attention, especially for long-term cases.

Due to the powerful self-attention mechanism that can flexibly attend to and connect tokens, Transformer~\cite{vaswani2017attention} architectures have been an appealing choice in a wide range of areas such as natural language processing (NLP) and image recognition. They have also been introduced to MTSF to capture the long-term time series dependency. 
MTSF Transformers are required to make a long-term forecast based on the observations of a multivariate time series, a vector consisting of $K (K > 1)$ time variables. % maybe define it before this sentence.
However, the direct application of the conventional Transformer architecture suffers from catastrophic performance degradation. LogSparse~\cite{li2019enhancing} and Informer~\cite{zhou2021informer} observed that it is partially due to the auto-regressive strategy employed by the Transformer decoder and replaced this strategy with a one forward operation. They proposed to provide the decoder with start tokens obtained from the input of the encoder and placeholder tokens that are as long as the target sequence. This approach significantly avoids cumulative errors caused by conventional auto-regressive strategy and is widely used in existing MTSF Transformers.

While the change of the decoder forward operation has become an essential step in the transition from the conventional Transformer architecture to the MTSF Transformer architecture, there are still some core configurations of the Transformer architecture that need further consideration. One of them is the tokenization strategy. For MTSF Transformer architectures, it seems natural and undoubted to treat the simultaneous data points as a token. A series of methods~\cite{li2019enhancing, zhou2021informer, wu2021autoformer, liu2021pyraformer, zhou2022fedformer} adopts this strategy in their Transformer architecture and achieves seemingly promising results. 
But what \textit{was} right does not mean it \textit{is} right. We show that this tokenization strategy is problematic both theoretically and experimentally.

From the theoretical perspective, the tokenization strategy widely used in MTSF Transformer architectures ignores the \textit{token uniformity} inductive bias. Dong et al.~\cite{rankCollapse2021} have discovered that self-attention-based models possess a strong inductive bias towards token uniformity. They theoretically prove that the pure self-attention mechanism without skip connections loses rank exponentially with depth, which makes the output converge fast to a rank-1 matrix. Although the skip connections ease the degree of rank loss, the trend toward token uniformity can still be observed in a wide range of Transformer models. This inductive bias forces the predicted tokens of MTSF Transformers to be consistent, resulting in over-smooth predictions. From the experimental perspective, to validate this theory, we utilize Euclidean distance to measure the similarity between tokens produced by a pretrained vanilla MTSF Transformer that employs the one forward operation. 
Compared with the ground truth similarity maps, the prediction similarity maps reflect that the tokens in a cycle lack of high-frequency variations, which causes the predicted sequence curves to present an over-smoothing state.

To tackle the root of the over-smoothing problem, we vary the tokenization strategy from \textit{Time Point based Tokenization (TPT)} to \textit{Time Variable based Tokenization (TVT)}.
% To get to the root of the  problem associated with this tokenization strategy, we propose to 
Technically, instead of applying the conventional strategy that treats the data points at the same time as a \textit{time point token}, we treat each time variable in the $K$-variate time series as a \textit{time variable token}. Therefore, the token uniformity inductive bias would act on the variable dimension rather than the time dimension. Compared with the TPT strategy, the TVT strategy can reduce the over-smoothing degree of predictions in the time dimension and enhance the associations between different variables in the variable dimension. It can be observed in the experiments that a vanilla TVT based Transformer is significantly superior to a TPT based Transformer. 

Furthermore, we have observed that both the self-attention and cross-attention mechanisms in the decoder of TVT based Transformers play little role in purposefully connecting tokens due to the limited information contained in the placeholders. Following the principle of simplicity, we replace the elaborately designed Transformer decoder with only one linear layer to output predictions. Surprisingly, this simple modification further improves the forecasting performance. Experiments on five widely-used benchmarks demonstrate that the final architecture even substantially outperforms the state-of-the-art Transformers in most cases.

To summarize, our key contributions are described as follows:
\begin{enumerate}
    \item We theoretically and experimentally point out that the tokenization strategy widely used in existing MTSF Transformers is flawed, which easily leads to an over-smoothing prediction.
    \item We propose a simple yet effective tokenization strategy for MTSF Transformers, and redesign the decoder and the embedding approach to match with this strategy. It is a new transformer paradigm for multivariate time series forecasting. The evolved architecture, which is simple enough to demonstrate the power of our core idea,  surprisingly beats all the state-of-the-art MTSF Transformers in most cases.
\end{enumerate}

\section{Related Work}

\subsection{Non-Transformer Models for Time Series Forecasting}
\par There are various methods designed for the multi-variate time series forecasting problem. Classical methods like ARIMA~\cite{box2015timeARIMA, box1968someARTMA} utilized temporal difference and moving average to forecast future points. Recently, neural network based methods have achieved promising results. Recurrent Neural Network (RNN) was introduced to model the temporal dependencies by hidden states~\cite{wen2017multi, rangapuram2018deep}. DeepAR~\cite{salinas2020deepar} combined auto-regressive methods and RNN to model the statistics of time series such as mean value and standard deviation. LSTNet~\cite{lai2018modeling} exploited Convolution Neural Network (CNN) and RNN to extract short-term local dependency and long-term global correlation. With the development of the attention mechanism, RNN-based models were improved to capture long-range dependency. However, these RNN-based models suffer from slow inference speed and error accumulation caused by the iterative step forward. Besides, gradient vanishing or explosion challenges the stability of the model. Temporal Convolution Network (TCN)~\cite{oord2016wavenet, borovykh2017conditional, bai2018empirical, sen2019think} based models are another family of conventional forecasting models, which use causal convolution to model temporal dependency. However, limited to the reception field of the kernel, extracted features still stay in the local stage and long-range dependency is hard to grasp.

\subsection{Transformers for Multi-variate Time Series Forecasting}
\label{sec:transformers-for-time-series-forecasting}
\par Recently, self-attention mechanism based Transformers~\cite{vaswani2017attention} show great power in modeling sequential data in a wide range of fields, such as natural language processing (NLP)~\cite{vaswani2017attention, devlin2018bert, radford2018improving}, audio processing~\cite{huang2018music} and even computer vision (CV)~\cite{dosovitskiy2020vit}. However, directly applying self-attention to long time series is computational prohibitive, %because the complexity is quadratic with the length $L$ of time series both in time and memory.
because the vanilla dynamic decoding process is still an auto-regressive manner. Decoder of Informer~\cite{zhou2021informer} takes input as the concatenation of start token and padding zeros/mean values, and directly forecasts all future points. This direct forecasting manner avoids auto-regressive manner and achieves faster speed. Following Transformers for multi-variate time series forecasting also adopt this direct forecasting manner. Besides, many variants of Transformer, such as FEDformer~\cite{zhou2022fedformer}, Autoformer~\cite{wu2021autoformer} and Prayformer~\cite{liu2021pyraformer}, et al., redesign the self-attention mechanism to reduce the complexity while preserving the capacity of modeling sequential data. Autoformer~\cite{wu2021autoformer} replaces conventional self-attention with an Auto-Correlation module based on the series periodicity, which conducts the function of dependency discovery and representation aggregation at the sub-series level and reduces complexity to $O(L \log L)$, where $L$ is the length of sequence. FEDformer~\cite{zhou2022fedformer} applies self-attention on frequency domain using Fourier transform~\cite{bracewell1986fourier} or wavelet transform~\cite{zhang2019wavelet}, and randomly selects frequency basis to compute the attention map, thus reducing the cost to $O(L)$. Prayformer~\cite{liu2021pyraformer} models the attention by a tree structure on different resolutions and reduces cost to $O(L)$. 

In general, their self-attention modules are all applied in the temporal dimension to realize time point mixing, and time variable mixing is realized by feed-forward networks. 
It seems to be a natural and unquestioned idea, but is it the most suitable strategy? Based on our analysis and experiments, time variable mixing by self-attention is flawed. Channel mixing realized by self-attention is more appropriate for most multi-variate time series, and the forecasting results are more accurate than FFN-based channel mixing.

\section{Definition of the MTSF Problem}
Given a $K$-variate ($K > 1$) time series $\mathbf{X}^t = \{\mathbf{x}^t_{t-L+1}, \mathbf{x}^t_{t-L+2}, \dots, \mathbf{x}^t_{t} \mid \mathbf{x}^t_{t-L+i} \in \mathbb{R}^K, i=1,2,\dots,L \}  \in \mathbb{R}^{K \times L}$ with a fixed observed window length $L$ at time $t$, multivariate time-series forecasting (MTSF) requires to accurately predict the consequent $K$-variate time series $\mathbf{Y}^t = \{\mathbf{y}^t_{t+1}, \mathbf{y}^t_{t+2}, \dots, \mathbf{y}^t_{t+H} \mid \mathbf{y}^t_{t+j} \in \mathbb{R}^K, j=1,2,\dots,H \}  \in \mathbb{R}^{K \times H}$ with a horizon $H$. In the challenging long-term cases, usually the horizon $H$ is comparable to or far more larger than the observed window length $L$.

\section{The Start Point of Evolution:\\ Vanilla MTSF Transformer}

We first provide a brief introduction on the conventional Transformer architecture, which is the foundation of the vanilla MTSF Transformer. Then we conclude three core configurations of the vanilla MTSF Transformer architecture in details. The overall architecture of the vanilla MTSF Transformer can be seen in the top of Fig.~\ref{fig:elv}. A PyTorch-like pseudo code can be found in Alg.~\ref{alg:vanilla_MTSF}.

\subsection{Conventional Transformer Architecture}
Transformer~\cite{vaswani2017attention} architecture has become a sizzling hot choice in a wide range of areas like natural language process (NLP) and computer vision (CV). It divides the input sequence into tokens with smaller granularity, and employs a flexible attention mechanism to connect related tokens and strengthen their correlations. Using an encoder-decoder structure, Transformer maps the input sequence into a compact implicit representation by its encoder, and maps this representation back to the target sequence by its decoder. The encoder receives its input and applies the self-attention mechanism to processes the input in parallel. The decoder employs a dynamic decoding strategy to predict results. It takes its previous outputs before time $t$ as input, and utilizes the cross-attention mechanism to combine the current implicit representation generated by the encoder to predict the output at time $t+1$. To make use of the order of the sequence, the input of the conventional Transformer is embedded by sine and cosine functions of different frequencies.

\subsection{Vanilla MTSF Transformer Architecture}
Except for the one forward operation utilized by the decoder, the core configurations of the vanilla MTSF Transformer are merely consistent with that of the conventional Transformer.
Despite the fact that a series of MTSF Transformers~\cite{li2019enhancing, zhou2021informer, wu2021autoformer, liu2021pyraformer, zhou2022fedformer} have designed different structures to reduce the complexity of the attention mechanism and obtain a more compact representation of a time series, generally there are three core configurations of the MTSF Transformer architectures that remain unchanged in the previous methods.

\paragraph{Tokenization Strategy}
In the field of multivariate time series forecasting, it seems straightforward and unquestioned to treat the data points at the same time as a token, which is formulized as:

\begin{equation}
    \label{eq:typical_tokenization}
    \mathbf{X}^t = \mathbf{X}^t_{TPT} = \{\mathbf{x}^t_{t-L+1}, \dots, \mathbf{x}^t_{t} \mid \mathbf{x}^t_{t-L+i} \in \mathbb{R}^K \} \in \mathbb{R}^{K \times L}
\end{equation}
where $i = 1, 2, \dots, L$. This Time Point based Tokenization (TPT) strategy considers a $K$-variate time series as a sequence of $K$-variate vectors indexed in time order, which are consistent with the tokenization approach in conventional Transformers.

\paragraph{Encoder-decoder Structure}
The encoder-decoder structure~\cite{cho2014learning} is firstly proposed for RNN, but has been universal among Transformer architectures. Current MTSF Transformers adopt this encoder-decoder structure. They employ a Transformer encoder to map the input time series into a fixed-length vector, and utilize a Transformer decoder to map the vector to the target output. Different from the decoder of the conventional Transformer, the vanilla MTSF Transformer decoder applies a one forward operation rather than a typical auto-regressive approach. The input of the decoder consists of start tokens $\mathbf{X}^t_{start}$ coming from the latter part of the observed time series, and the placeholder tokens $\mathbf{X}^t_{0}$ filled with the same padding value (e.g., 0 or the mean value of the input series) for the target sequence. It’s formulized as follows:

\begin{equation}
    \mathbf{X}^t_{dec} = \mathrm{Concat}(\mathbf{X}^t_{start}, \mathbf{X}^t_{0}) \in \mathbb{R}^{d_{model} \times (L_{start} + H)}
\end{equation}
where $d_{model}$ means the model dimension, $L_{start}$ means the length of the start tokens, and $H$ refers to the horizon.

\paragraph{Input Representation}
Conventional Transformers only use sinusoidal and cosinusoidal positional embeddings, which are formulized as follows:
\begin{align}
    \mathrm{PE}_{(i, 2j)} &= \sin (i / (2L)^{2j / d_{model}}) \\
    \mathrm{PE}_{(i, 2j+1)} &= \cos (i / (2L)^{2j / d_{model}})
\end{align}
where $j = 0, 1, \dots, \lfloor d_{model} / 2 \rfloor $.
The existing MTSF Transformers further add hierarchical time stamp groups (hour, week, month and year) to capture the global context. Each group of global time stamps is mapped to a learnable stamp embeddings by a function $\mathrm{SE}$. 
To map the input $K$-variate time series into the model dimension $d_{model}$, the original scalars are projected from $K$-dim into $d_{model}$-dim by 1-D convolutional filters $\mathrm{Conv}$ (kernel width=3, stride=1). Therefore, the feeding vector of the vanilla MTSF Transformer is: 

\begin{equation}
    \mathbf{X}^t_{\mathrm{feed}[i]} = \mathrm{Conv}(\mathbf{X}^t_{i-1:i+2}) + \mathrm{PE}_{i} + \mathrm{SE}(\mathrm{Stamp}_{i})
\end{equation}
where $i = 1, 2, \dots, L $. It is noticed that since some methods associate tokens with auto-correlation mechanism~\cite{wu2021autoformer} or Fourier Transform~\cite{zhou2022fedformer} which inherently contains local context, there is no need for them to add positional embeddings.

\begin{figure*}[!ht]
    \centering
    \includegraphics[width=1.\linewidth]{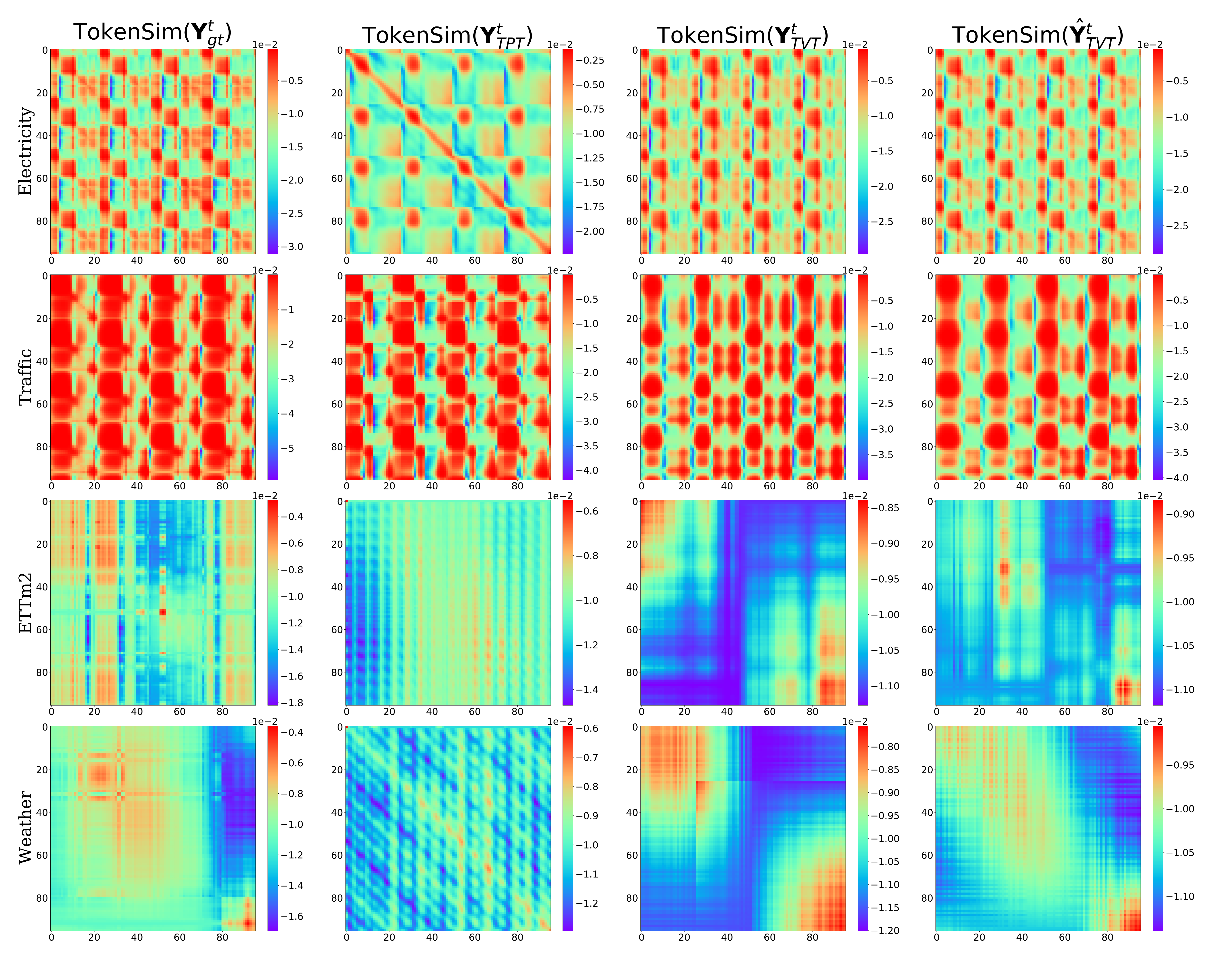}
    \caption{Similarity heat maps derived from the ground truth $\mathbf{Y}_{gt}^t$, predicted series $\mathbf{Y}^t_{TPT}$ from a pretrained TPT based Transformer, $\mathbf{Y}^t_{TVT}$ from a pretrained TVT based Transformer with a Transformer decoder and $\hat{\mathbf{Y}}^t_{TVT}$ from a pretrained TVT based Transformer with a linear layer decoder. They are derived from test datasets of electricity, traffic, ETTm2 and weather. The predicted horizon $H$ is set to be 96. It is easy to find that $\mathrm{TokenSim}(\mathbf{Y}^t_{TVT})$ and  $\mathrm{TokenSim}(\hat{\mathbf{Y}}^t_{TVT})$ are more similar to $\mathrm{TokenSim}(\mathbf{Y}_{gt}^t)$.}
    \label{fig:SimMap}
\end{figure*}

\begin{figure*}[!ht]
    \centering
    \includegraphics[width=1.\linewidth]{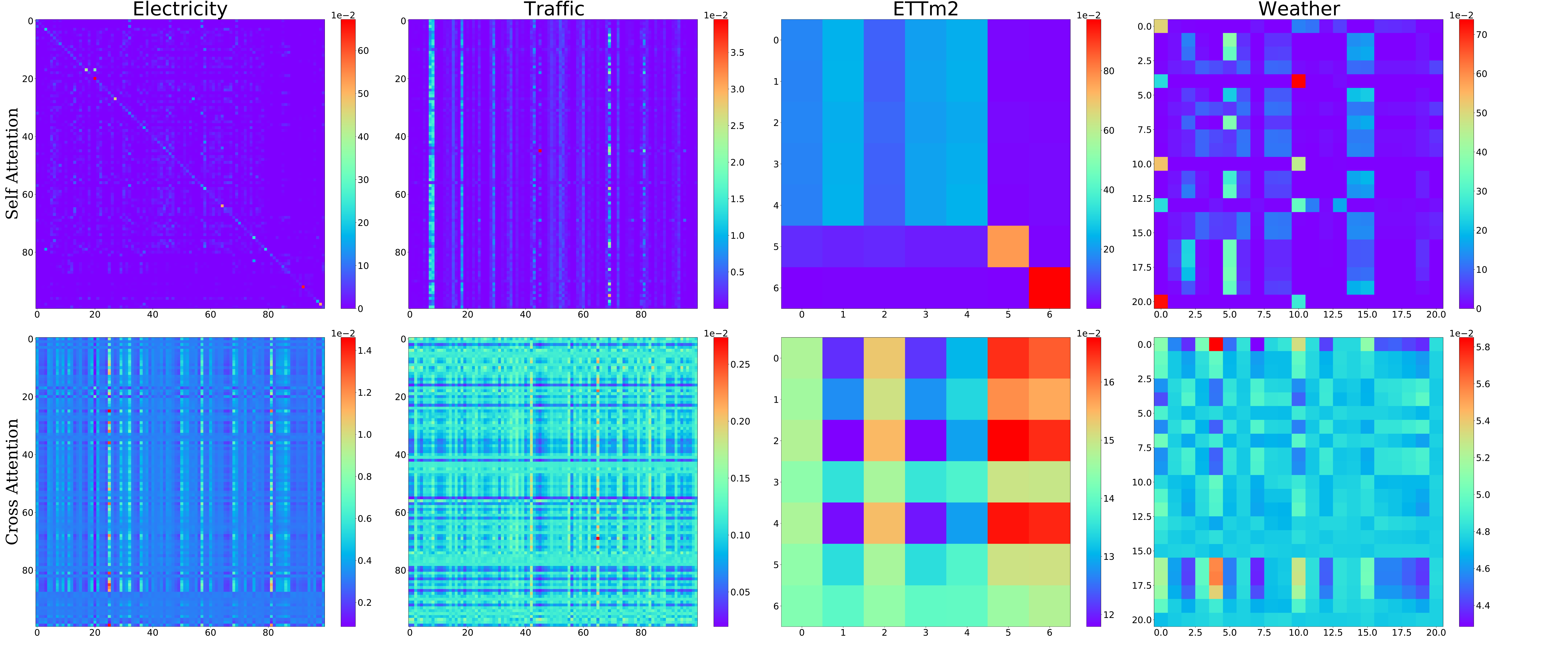}
    \caption{Self-attention and cross-attention score maps from the one-layer Transformer decoder of TVT Transformers, which are pretrained in different datasets (electricity, traffic, ETTm2 and weather) and tested on the test sets. The attention maps are very disorganized, which means both the self-attention and cross-attention module do not focus on the effective information of the time variables.}
    \label{fig:AttDecoder}
\end{figure*}

\section{The Progressive Process of Evolution:\\Time Point Tokenization (TPT) Vs Time Variable Tokenization (TVT)}

In this section we first describe the issue in the current tokenization strategy for vanilla MTSF Transformers in detail, and then propose a new tokenization strategy to deal with it. Next, we go further to improve the Transformer architecture by ablating the decoder and directly applying a simple linear layer to the output of the encoder. We also update the embedding strategy to match with the final architecture.

\subsection{The Devil Is in the Tokenization Strategy}
According to Dong et al.~\cite{rankCollapse2021}, the pure self-attention mechanism naturally forces the output tokens to be biased towards a \textit{token uniformity} state, in which the rank of the output matrix collapses doubly exponentially. 

For a self-attention network $\mathrm{SAN}$ with depth $L$ and $H$ heads and without skip connection, supposing the norm of the weight matrix $\Vert \mathbf{W}^l_{QK, h} \Vert_1 \Vert \mathbf{W}_h^l \Vert_{1, \infty}$ has an upper boundary $\beta$ for any $h \in \{1, 2, \dots, H \}$ and $l \in \{1, 2, \dots, L  \}$, it is given that
\begin{equation}\label{eq:rankCollapse}
    \Vert \mathrm{RES}(\mathrm{SAN}(\mathbf{X})) \Vert_{1, \infty} \leq (\frac{4\beta H}{\sqrt{d_{qk}}})^{\frac{3^L-1}{2}} \Vert \mathrm{RES}(\mathbf{X}) \Vert_{1, \infty}^{3^L},
\end{equation}
where $\mathbf{X} \in \mathbb{R}^{N_{token} \times D_{model}}$ represents the input of the self-attention network, and $\mathrm{SAN}(\mathbf{X})$ represents the output. The residual of $\mathbf{X}$ is defined as $\mathrm{RES}(\mathbf{X}) = \mathbf{X} - \textbf{1} \mathbf{x}^{\top}$, where $\mathbf{x} = \arg\min_{\mathbf{x}} \Vert \mathbf{X} - \textbf{1} \mathbf{x}^{\top} \Vert$.
The $\mathbf{L}_1, \mathbf{L}_{\infty}$-composite norm of a matrix $\mathbf{X}$ is defined as $\Vert \mathbf{X} \Vert_{1, \infty} = \sqrt{\Vert \mathbf{X} \Vert_{1} \Vert \mathbf{X} \Vert_{\infty}}$. 

Eq.~\ref{eq:rankCollapse} indicates that, when subtracting a rank-$1$ matrix from the input and the output of the pure self-attention network, the residual of the output is exponentially smaller than the residual of the input, which represents a rank collapse of $\mathbf{X}$. The tokens of the output $\mathrm{SAN}(\mathbf{X})$ get more similar to each other, and present a token uniformity state. Although the strict derivation of the formula holds only in the case of pure self-attention without skip connections, it is observed that the rank collapse of the output matrix also occurs on a wide range of Transformer models in practice.

Everything has its pros and cons. In some scenarios, the token uniformity inductive bias of Transformers can connect related tokens and strengthen the associations among them to some extent. However, we find this inductive bias is a performance bottleneck for the vanilla MTSF Transformer. A vanilla MTSF Transformer employs a tokenization strategy which we call \textit{Time Point Tokenization (TPT)}. It treats data points at the same time as a token, which is formulized by Eq.~\ref{eq:typical_tokenization}. Due to the token uniformity inductive bias of Transformers, tokens at different time are forced to be uniform, which makes the predicted time series over-smooth and lacks of high frequency variations.

To verify this hypothesis, we propose to measure the similarity between two predicted time point tokens by Euclidean distance, which is also the criterion function for model training. For the predicted result $\mathbf{Y}^t = \{\mathbf{y}^t_{t+1}, \mathbf{y}^t_{t+2}, \dots, \mathbf{y}^t_{t+H} \mid \mathbf{y}^t_{t+j} \in \mathbb{R}^K, j=1,2,\dots,H \}  \in \mathbb{R}^{K \times H}$ from a TPT based Transformer at time $t$, we first define the Euclidean distance similarity matrix $\mathbf{E}^t_{tokens}$ as:
% \begin{equation}
%     \mathbf{E}^t_{tokens} = ( \Vert \mathbf{y}^t_{t+i} - \mathbf{y}^t_{t+j} \Vert_2 )_{i, j} \in \mathbb{R}^{H \times H} 
% \end{equation}
\begin{equation}
    \begin{pmatrix}
        \Vert \mathbf{y}^t_{t+1} - \mathbf{y}^t_{t+1} \Vert_2 ,& \cdots ,& \Vert \mathbf{y}^t_{t+1} - \mathbf{y}^t_{t+H} \Vert_2\\
        \Vert \mathbf{y}^t_{t+2} - \mathbf{y}^t_{t+1} \Vert_2 ,& \cdots ,& \Vert \mathbf{y}^t_{t+2} - \mathbf{y}^t_{t+H} \Vert_2\\
        \vdots &  \ddots & \vdots\\
        \Vert \mathbf{y}^t_{t+H} - \mathbf{y}^t_{t+1} \Vert_2 ,& \cdots ,& \Vert \mathbf{y}^t_{t+H} - \mathbf{y}^t_{t+H} \Vert_2\\
    \end{pmatrix}
\in \mathbb{R}^{H \times H} 
\end{equation}
$\Vert \cdot \Vert_{2}$ means the Euclidean norm. Then the similarity between predicted time point tokens at time $t$ can be calculated by a function $\mathrm{TokenSim}$:
\begin{equation}
    \mathrm{TokenSim}(\mathbf{Y}^t) = -\mathrm{softmax}(\mathbf{E}^t_{tokens}) \in \mathbb{R}^{H \times H} 
\end{equation}
The minus sign indicates that a higher result value (deeper red) corresponds to a higher similarity between the predicted time point token pairs.

We utilize the $\mathrm{TokenSim}$ function to calculate the similarity between time point tokens of the time series $\mathbf{Y}^t_{TPT}$ predicted by a pretrained TPT based Transformer and the ground truth $\mathbf{Y}_{gt}^t$ at different datasets. We present the results derived from electricity, traffic, ETTm2 and weather datasets, which have relatively strong cyclical nature. The results are visualized in Fig.~\ref{fig:SimMap}. The predicted horizon $H$ is set to be 96. The heat maps in the first and second column come from the ground truth $\mathbf{Y}_{gt}^t$ and the predicted time series $\mathbf{Y}^t_{TPT}$ respectively. 

All the heat maps can be viewed as consisting of $4 \times 4$ similar blocks. The size of each block is $24 \times 24$, and 24 (hours) represents the cycle of the time series. Compared with the blocks in heat maps of $\mathbf{Y}_{gt}^t$ in the first column, the blocks in heat maps of $\mathbf{Y}^t_{TPT}$ in the second column are distinctly smoother and lack of high frequency details, which indicates the predicted time series suffer from being over-smoothing and missing high frequency components.

%The periodically varying heat maps indicate to what extend the tokens at a certain interval of time are relatively consistent, which means to a certain degree of periodicity of the time series. It can be easily observed that the ground truth heat maps in the first column have a noticeable cycle length of 24 (hours), while there exist different details within different cycles. By contrast, the heat maps derived from the TPT based Transformer in the second column present a over-head periodicity. The similarity values within a cycle are over-smooth and relatively highly consistent with the values within the different cycles. There are limited distinct details 

% Therefore, there exist some degree of cyclicality in their ground truth similarity maps. However, all the similarity maps of the predicted time series present remarkable cyclicality which is more significant than that of the respective ground truth. The cyclicality of the similarity maps reflects the similarity between tokens at a certain time point interval. With higher degree of cyclicality in the similarity maps, the outputs of TPT transformers are also with higher periodicity, which suffers from over-smoothing and lack of high frequency components.

To reduce the negative impacts brought by the \textit{token uniformity} inductive bias, one way is to apply a more effective time-series representation to aggregate the tokens at different time points and thus avoid a over-smoothing prediction, such like auto-correlation~\cite{wu2021autoformer} or Fourier transform~\cite{zhou2022fedformer}. However, we claim that it is more straightforward to tackle the problem at its root by just changing the tokenization strategy of Transformers. While \textit{token uniformity} is harmful to the TPT based Transformer architectures, it is beneficial to strengthen the similarity between different variables. The self-attention mechanism can flexibly compensate each variable with information from its covariance, which the feed-forward networks from TPT based Transformers are incompetent at. Furthermore, when treating a variable rather than some data points at a time as a token, it significantly eases the over-smoothing phenomenon. We call this tokenization strategy as \textit{Time Variable Tokenization (TVT)}. It can be easily observed that the TVT based transformer significantly overcomes the over-smoothing problem, and produces a time series with high frequency details. 
As shown in Fig.~\ref{fig:SimMap}, the third columns and fourth columns are derived from the output $\mathbf{Y}^t_{TVT}$ and $\hat{\mathbf{Y}}^t_{TVT}$ produced by two types of TVT based Transformers, respectively. The similar cycle blocks contain more high frequency details that are consistent with the blocks from the ground truth, which indicates the ease of over-smoothing phenomenon.
By reducing the over-smoothing of TPT based Transformers and connect related variables more effectively, the TVT based Transformer achieves a remarkable MTSF performance in different datasets and substantially beats the vanilla TPT based transformer.

\subsection{Is the Transformer Decoder Still Necessary?}
We go further to analyze and optimize the TVT based Transformer architecture. The input $K$-variate time series $\mathbf{X}^t \in \mathbb{R}^{K \times L}$ are now tokenized to be $K$ variable tokens: $\{ \mathbf{x}^t_{1}, \mathbf{x}^t_{2}, \dots, \mathbf{x}^t_{K} \mid \mathbf{x}^t_i \in \mathbb{R}^{L}, i = 1, 2, \dots, K \}$. 
% In despite the fact that the input time variable tokens have no obvious local \textit{positional} correlations, such like temporal correlations in tokens from NLP Transformers and spatial correlations in tokens from Vision Transformers, each token pair from the TVT based Transformer has a certain degree of dependency due to the prior correlations between 
% the tokens. The prior correlations can be similar electricity consumption habits, similar traffic flows, etc.
% This prior correlations are inherently contained in the fixed \textit{position} of each time variable. 
Since the input variable tokens are indexed by a fixed channel order, there is no need to employ a Transformer decoder to sample associated tokens and merge their information again, which is already handled by the Transformer encoder. Moreover, since the input tokens of the transformer decoder mainly contain the information from time stamps, it is meaningless to calculate the similarity between tokens by either the self-attention mechanism or the cross-attention mechanism. Fig.~\ref{fig:AttDecoder} presents self-attention and cross-attention score maps derived from the one-layer Transformer decoder of TVT based Transformers pretrained in different datasets. The attention maps are very disorganized, which means both the self-attention and cross-attention module do not focus on the effective information of the time variables.

% \begin{figure*}
%     \centering
%     \includegraphics[width=0.9\linewidth]{pics/decoder_channelTokensTransformer.pdf}
%     \caption{Self attention scores and cross attention scores from TVT based Transformers pretrained in different datasets. More visualisation results are provided in the supplementary materials.}
%     \label{fig:AttTVT}
% \end{figure*}

Based on the above observations, we propose to discard the meaningless Transformer decoder. Following the principle of simplicity, we only apply a simple linear layer on top of the Transformer encoder to support length variations from the model dimension $d_{model}$ to the horizon $H$. Surprisingly, this simple architecture works well and successfully exceeded the architecture with Transformer decoder.

\subsection{Adaptive Embedding Approaches}
Vanilla MTSF Transformers need amounts of global and local embeddings to help capture the temporal correlations between tokens. This is because the self-attention mechanism is parallel, and there is no sequential relationship between time point-based tokens. When the tokenization strategy is changed from TPT to TVT, the meaning of the ``positional” relationship between tokens also varies. 
Different from the temporal correlations between tokens in TPT based Transformers, each token pair from the TVT based Transformer has a certain degree of dependency due to the prior correlations between the tokens. These prior correlations can be similar consumers' electricity consumption habits; similar traffic flows on different roads, etc.
Therefore, the ``positional” correlations between tokens in TVT based Transformers are inherently fixed by order of tokens. 
When applying TVT to a transformer, we can abandon the sinusoidal and cosinusoidal positional embedding strategy used for vanilla Transformers.

\begin{figure*}[ht]
    \centering
    \includegraphics[width= 0.9\linewidth]{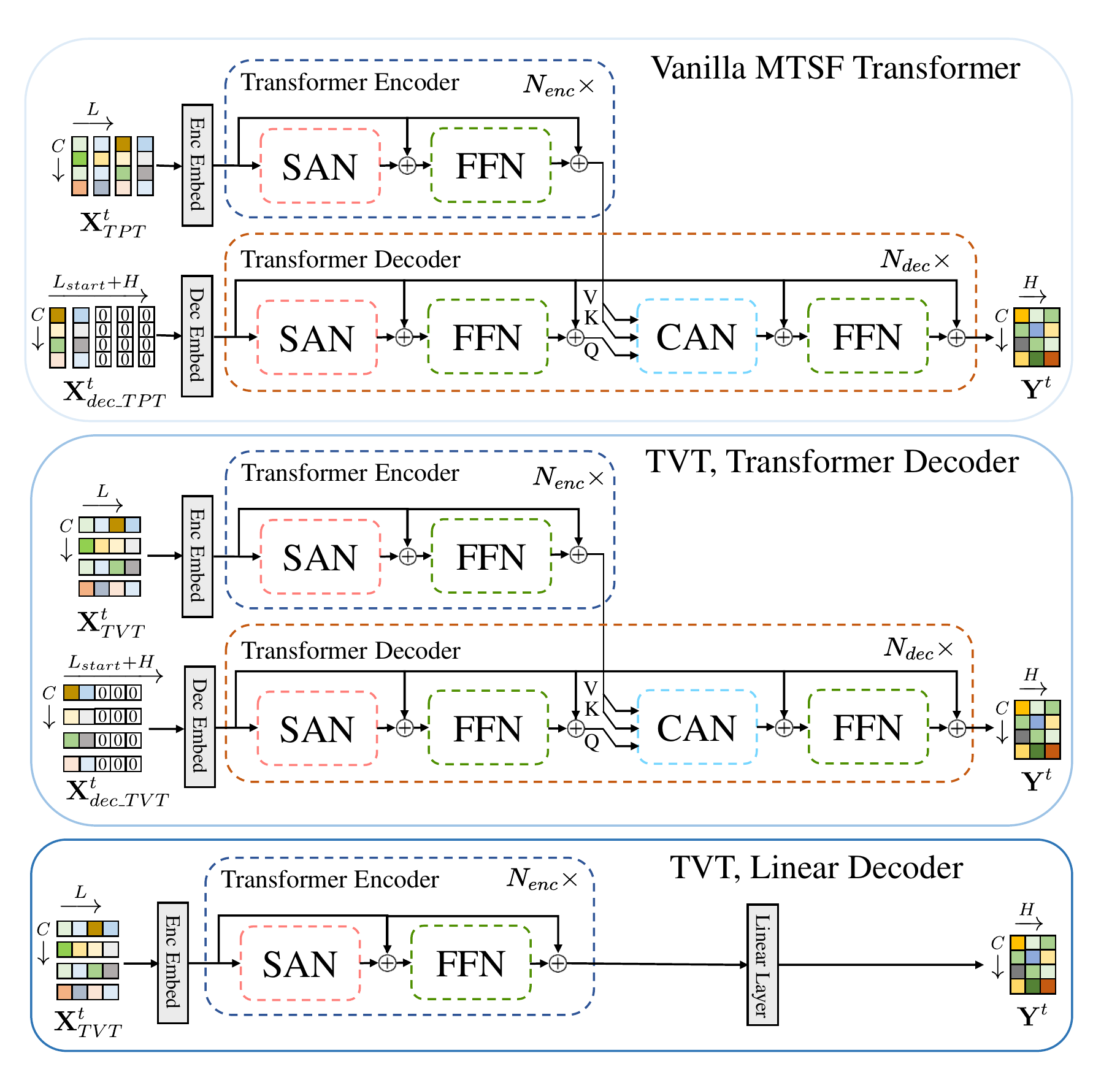}
    \caption{Top to bottom: vanilla MTSF Transformer, TVT based Transformer with a Transformer decoder, and TVT based Transformer with a linear decoder. The vanilla MTSF Transformer utilizes Time Point Tokenization (TPT) and treats the data points at the same time as a token. By contrast, our Time Variable Tokenization (TVT) based solutions treat each time variable as a token.}
    \label{fig:elv}
\end{figure*}

\begin{algorithm}[ht]
\caption{{Code for vanilla MTSF Transformer (TPT based) (PyTorch-like)}}
\label{alg:vanilla_MTSF}
\definecolor{codeblue}{rgb}{0.25,0.5,0.25}
\lstset{
	backgroundcolor=\color{white},
	basicstyle=\fontsize{7.2pt}{7.2pt}\ttfamily\selectfont,
	columns=fullflexible,
	breaklines=true,
	captionpos=b,
	commentstyle=\fontsize{7.2pt}{7.2pt}\color{codeblue},
	keywordstyle=\fontsize{7.2pt}{7.2pt},
	%  frame=tb,
}
\begin{lstlisting}[language=python]
# B: batch size, C: channel/variable, L: observed window length, H: horizon length, D: d_model
# x: observed time series of shape (B, L, C)
# x_mark: corresponding time stamps of shape (B, L, S)
# y: Concatenation of starters and placeholders of shape (B, L_start + H, C)
# y_mark: corresponding time stamps of shape (B, L_start + H, S)


#<--------------- initialization --------------->
# Initialize Transformer Encoder
encoder_layer = nn.TransformerEncoderLayer(D, nhead, batch_first=True)
self.transformer_encoder = nn.TransformerEncoder(encoder_layer, num_enc)

# Initialize Transformer Decoder
decoder_layer = nn.TransformerDecoderLayer(D, nhead, batch_first=True)
self.transformer_decoder = nn.TransformerDecoder(decoder_layer, num_dec)

# Project d_model into output channel
self.proj = nn.Linear(D, C)

# Input Representation
self.enc_embed, self.dec_embed = get_embed(args.embed_type)


#<--------------- code in forward --------------->
def forward(self, x, x_mark, y, y_mark):
    enc_in = self.enc_embed(x, x_mark)
    enc_out = self.transformer_encoder(enc_in)
    
    dec_in = self.dec_embed(y, y_mark)
    dec_out = self.transformer_decoder(dec_in, enc_out)
    
    pred = dec_out[:, -H:, :]
    pred = self.proj(pred) # shape (B, H, C)
    return pred
\end{lstlisting}
\end{algorithm}

\begin{algorithm}[t]
\caption{{Code for TVT Transformer with a Transformer decoder (PyTorch-like)}}
\label{alg:TVT_Trans}
\definecolor{codeblue}{rgb}{0.25,0.5,0.25}
\lstset{
	backgroundcolor=\color{white},
	basicstyle=\fontsize{7.2pt}{7.2pt}\ttfamily\selectfont,
	columns=fullflexible,
	breaklines=true,
	captionpos=b,
	commentstyle=\fontsize{7.2pt}{7.2pt}\color{codeblue},
	keywordstyle=\fontsize{7.2pt}{7.2pt},
	%  frame=tb,
}
\begin{lstlisting}[language=python]
# B: batch size, C: channel/variable, L: observed window length, H: horizon length, D: d_model
# x: observed time series of shape (B, L, C)
# x_mark: corresponding time stamps of shape (B, L, S)
# y: Concatenation of starters and placeholders of shape (B, L_start + H, C)
# y_mark: corresponding time stamps of shape (B, L_start + H, S)


#<--------------- initialization --------------->
# Initialize Transformer Encoder
encoder_layer = nn.TransformerEncoderLayer(C, nhead, batch_first=True)
self.transformer_encoder = nn.TransformerEncoder(encoder_layer, num_enc)

# Initialize Transformer Decoder
decoder_layer = nn.TransformerDecoderLayer(C, nhead, batch_first=True)
self.transformer_decoder = nn.TransformerDecoder(decoder_layer, num_dec)

# Project d_model into horizon length
self.proj = nn.Linear(D, H)

# Input Representation
self.enc_embed, self.dec_embed = get_embed(args.embed_type)


#<--------------- code in forward --------------->
def forward(self, x, x_mark, y, y_mark):
    enc_in = self.enc_embed(x, x_mark)
    enc_in = enc_in.permute(0, 2, 1) # shape (B, D, C) -> (B, C, D)
    enc_out = self.transformer_encoder(enc_in)
    
    dec_in = self.dec_embed(y, y_mark)
    dec_in = dec_in.permute(0, 2, 1) # shape (B, D, C) -> (B, C, D)
    dec_out = self.transformer_decoder(dec_in, enc_out)
    
    pred = self.proj(dec_out) # shape (B, C, D) -> (B, C, H)
    pred = pred.permute(0, 2, 1) # shape (B, C, H) -> (B, H, C)
    return pred
\end{lstlisting}
\end{algorithm}

\begin{algorithm}[t]
\caption{{Code for TVT Transformer with a Linear decoder (PyTorch-like)}}
\label{alg:TVT_Linear}
\definecolor{codeblue}{rgb}{0.25,0.5,0.25}
\lstset{
	backgroundcolor=\color{white},
	basicstyle=\fontsize{7.2pt}{7.2pt}\ttfamily\selectfont,
	columns=fullflexible,
	breaklines=true,
	captionpos=b,
	commentstyle=\fontsize{7.2pt}{7.2pt}\color{codeblue},
	keywordstyle=\fontsize{7.2pt}{7.2pt},
	%  frame=tb,
}
\begin{lstlisting}[language=python]
# B: batch size, C: channel/variable, L: observed window length, H: horizon length, D: d_model
# x: observed time series of shape (B, L, C)
# x_mark: corresponding time stamps of shape (B, L, S)
# y: Concatenation of starters and placeholders of shape (B, L_start + H, C)
# y_mark: corresponding time stamps of shape (B, L_start + H, S)


#<--------------- initialization --------------->
# Initialize Transformer Encoder
encoder_layer = nn.TransformerEncoderLayer(C, nhead, batch_first=True)
self.transformer_encoder = nn.TransformerEncoder(encoder_layer, num_enc)

# Initialize Linear Decoder
self.linear_decoder = nn.Linear(D, H)

# Input Representation
self.enc_embed, _ = get_embed(args.embed_type)


#<--------------- code in forward --------------->
def forward(self, x, x_mark, y, y_mark):
    enc_in = self.enc_embed(x, x_mark)
    enc_in = enc_in.permute(0, 2, 1) # shape (B, D, C) -> (B, C, D)
    enc_out = self.transformer_encoder(enc_in)
    
    pred = self.linear_decoder(enc_out)
    
    pred = pred.permute(0, 2, 1) # shape (B, C, H) -> (B, H, C)
    return pred
\end{lstlisting}
\end{algorithm}

\section{The Outcome of Evolution:\\TVT Based MTSF Transformer}

Finally we provide an outcome of the evolved TVT Transformer architecture. An overview of the two TVT based Transformer architectures is shown in the second and third rows of Fig.~\ref{fig:elv}. A PyTorch-like pseudo code can be found in Alg.~\ref{alg:TVT_Trans} and Alg.~\ref{alg:TVT_Linear}.

\paragraph{Tokenization Strategy}
The observed $K$-variate time series $\mathbf{X}^t \in \mathbb{R}^{K \times L}$ at time $t$ is split into $K$ tokens along the variable dimension, which is formulized as:
\begin{equation}
    \mathbf{X}^t = \mathbf{X}_{TVT}^t =  \{ \mathbf{x}^t_{1}, \mathbf{x}^t_{2}, \dots, \mathbf{x}^t_{K} \mid \mathbf{x}^t_i \in \mathbb{R}^{L} \},
\end{equation}
where $i = 1, 2, \dots, K$.
Then we apply a learnable linear projection function $\mathrm{E}_{pre} \in \mathbb{R}^{L \times D}$ to map $\mathbf{X}_{TVT}^t$ into $D$ dimensional time variable embeddings.
\begin{equation}
    \mathbf{Z}^t_{enc\_in} = \mathbf{X}_{TVT}^t \mathrm{E}_{pre} =  \{ \mathbf{x}^t_1 \mathrm{E}_{pre}, \mathbf{x}^t_2 \mathrm{E}_{pre}, \dots, \mathbf{x}^t_K \mathrm{E}_{pre} \}
\end{equation}

\paragraph{Encoder-decoder Structure}
The encoder of TVT based MTSF Transformers is consistent with that from conventional Transformers and vanilla MTSF Transformers. It consists of optional layers $N_{enc}$ each containing a multi-head self-attention module and a feed-forward network. Layer normalization is employed before each sub-layer, and the residual connection is after. The feed-forward network contains two linear layers with a GELU non-linearity.
The Transformer decoder consists of alternating layers $N_{dec}$ each containing a multi-head self-attention module, a multi-head cross-attention module and two feed-forward networks. We adopt the one forward operation from vanilla MTSF Transformer decoders. The input of the TVT based Transformer decoder is:
\begin{equation}
    \mathbf{X}^t_{dec} = \mathrm{Concat}(\mathbf{X}^t_{start}, \mathbf{X}^t_{0}) \in \mathbb{R}^{K \times (L_{start} + H)}
\end{equation}
where $\mathbf{X}^t_{start}$ comes from the latter piece of $\mathbf{X}^t$ with a length $L_{start}$, and $\mathbf{X}^t_{0}$ is placeholders filled with the same padding value 0 and is as long as the target sequence. Finally, a linear projection $\mathrm{E}_{post} \in \mathbb{R}^{D \times H}$ is utilized to project the tokens from $D$-dim to $H$-dim.
\begin{equation}
    \mathbf{Y}^t = \mathbf{Z}^t_{dec\_out} \mathrm{E}_{post} =  \{ \mathbf{z}^t_1 \mathrm{E}_{post}, \mathbf{z}^t_2 \mathrm{E}_{post}, \dots, \mathbf{z}^t_K \mathrm{E}_{post} \} 
\end{equation}
The Transformer decoder can further be replaced by a simple linear layer $\mathrm{E}_{fc} \in \mathbb{R}^{D \times H}$ that directly maps the output of the encoder to the target sequence.
\begin{equation}
    \mathbf{Y}^t = \mathbf{Z}^t_{enc\_out} \mathrm{E}_{fc} =  \{ \mathbf{z}^t_1 \mathrm{E}_{fc}, \mathbf{z}^t_2 \mathrm{E}_{fc}, \dots, \mathbf{z}^t_K \mathrm{E}_{fc} \} 
\end{equation}

\paragraph{Input Representation}
Due to the fully-connected layers in TVT based Transformers play a role in positional encoding, there is no need to add the sinusoidal and cosinusoidal positional embeddings. Therefore, only the time stamp groups $\mathrm{SE}$ are worth considering. We conduct experiments that are with or without the time stamp groups embeddings.

\begin{table*}[!ht]
    \centering
    \caption{Multivariate long-term series forecasting results among different Transformer-based models on five datasets with input length $I=96$ and forecasting length $O \in \{96,192,336,720\}$. Lower MSE and MAE indicate better performance. The best results are highlighted in \textbf{bold} and the second best results are highlighted with \underline{underline}.}
    \adjustbox{width=1.0\linewidth}{
        \begin{tabular}{c|c|cccc|cccc|cccc|cccc|cccc}
            \toprule
            \multirow{2}{*}{Methods} & \multirow{2}{*}{Metric} &\multicolumn{4}{c|}{ETTm2}&\multicolumn{4}{c|}{Electricity}&\multicolumn{4}{c|}{Exchange}&\multicolumn{4}{c|}{Traffic}&\multicolumn{4}{c}{Weather}\\
            & & 96 & 192 & 336 & 720 & 96 & 192 & 336 & 720 & 96 & 192 & 336 & 720 & 96 & 192 & 336 & 720 & 96 & 192 & 336 & 720\\
            \midrule

%%%%%%%%%%%%%%%%%%%%%%%%%%%%%%%%%%%%%%%%% paste start from here~
\multirow{2}{*}{Ours} &MSE &\textbf{0.181} &\textbf{0.249} &\textbf{0.324} &0.485 &\textbf{0.157} &\textbf{0.167} &\textbf{0.183} &\textbf{0.213} &\textbf{0.085} &\textbf{0.197} &\textbf{0.354} &\textbf{0.846} &\textbf{0.482} &\textbf{0.491} &\textbf{0.505} &\textbf{0.543} &\textbf{0.172} &\textbf{0.216} &\textbf{0.263} &\textbf{0.323} \\
 &MAE &\textbf{0.269} &\textbf{0.324} &\underline{0.381} &0.489 &\textbf{0.253} &\textbf{0.265} &\textbf{0.283} &\textbf{0.314} &\textbf{0.212} &\textbf{0.340} &\textbf{0.460} &\textbf{0.716} &\textbf{0.304} &\textbf{0.309} &\textbf{0.319} &\textbf{0.337} &\textbf{0.238} &\textbf{0.277} &\textbf{0.317} &\textbf{0.360} \\
\midrule
\multirow{2}{*}{FEDformer} &MSE &\underline{0.190} &\underline{0.257} &\underline{0.326} &\textbf{0.433} &\underline{0.188} &\underline{0.197} &\underline{0.214} &\underline{0.244} &\underline{0.138} &0.277 &\underline{0.448} &1.155 &\underline{0.576} &\underline{0.611} &\underline{0.620} &\underline{0.632} &\underline{0.254} &\underline{0.286} &\underline{0.343} &\underline{0.415} \\
 &MAE &\underline{0.284} &\underline{0.325} &\textbf{0.364} &\textbf{0.425} &\underline{0.303} &\underline{0.311} &\underline{0.328} &\underline{0.353} &\underline{0.267} &0.384 &\underline{0.493} &0.823 &\underline{0.359} &0.380 &0.381 &\underline{0.385} &0.340 &\underline{0.349} &\underline{0.382} &\underline{0.426} \\
\midrule
\multirow{2}{*}{Autoformer} &MSE &0.224 &0.277 &0.400 &\underline{0.447} &0.204 &0.233 &0.250 &0.273 &0.143 &\underline{0.272} &0.470 &\underline{1.098} &0.650 &0.616 &0.641 &0.675 &0.261 &0.335 &0.345 &0.431 \\
 &MAE &0.307 &0.334 &0.405 &\underline{0.432} &0.319 &0.336 &0.352 &0.375 &0.274 &\underline{0.381} &0.512 &\underline{0.813} &0.418 &0.386 &0.397 &0.419 &\underline{0.333} &0.388 &0.384 &0.433 \\
\midrule
\multirow{2}{*}{Pyraformer} &MSE &0.459 &0.536 &1.179 &3.748 &0.389 &0.378 &0.380 &0.375 &1.762 &1.838 &1.932 &2.092 &0.867 &0.868 &0.878 &0.892 &0.580 &0.726 &0.997 &1.433 \\
 &MAE &0.531 &0.555 &0.838 &1.441 &0.449 &0.442 &0.446 &0.444 &1.108 &1.135 &1.165 &1.207 &0.467 &0.465 &0.467 &0.470 &0.537 &0.617 &0.760 &0.938 \\
\midrule
\multirow{2}{*}{Informer} &MSE &0.398 &0.814 &1.467 &3.890 &0.319 &0.346 &0.346 &0.400 &0.944 &1.245 &1.793 &2.918 &0.746 &0.750 &0.849 &1.000 &0.384 &0.510 &0.812 &1.078 \\
 &MAE &0.487 &0.709 &0.929 &1.461 &0.406 &0.430 &0.430 &0.459 &0.773 &0.882 &1.072 &1.412 &0.415 &0.422 &0.478 &0.560 &0.446 &0.505 &0.650 &0.761 \\
\midrule
\multirow{2}{*}{Reformer} &MSE &0.686 &1.363 &2.021 &3.083 &0.300 &0.329 &0.344 &0.394 &0.907 &1.434 &1.846 &1.907 &0.694 &0.686 &0.689 &0.706 &0.346 &0.490 &0.766 &0.761 \\
 &MAE &0.627 &0.860 &1.053 &1.333 &0.401 &0.421 &0.430 &0.472 &0.767 &0.968 &1.119 &1.173 &0.387 &\underline{0.374} &\underline{0.375} &0.387 &0.391 &0.501 &0.636 &0.642 \\
\bottomrule
%%%%%%%%%%%%%%%%%%%%%%%%%%%%%%%%%%%%%%%%%%%%%%%%%%%%%%%%%%%%%%%% paste end here~
            
        \end{tabular}
    }
    \label{tab:formers-cmp}
\end{table*}
\begin{table*}[!ht]
    \centering
    \caption{Multivariate long-term series forecasting comparison among evolving and evolved Transformer models on five datasets with input length $I=96$ and forecasting length $O \in \{96,192,336,720\}$. Lower MSE and MAE indicate better performance. The best results are highlighted in \textbf{bold} and the second best results are highlighted with \underline{underline}.}
    \adjustbox{width=1.0\linewidth}{
        \begin{tabular}{c|c|cccc|cccc|cccc|cccc|cccc}
            \toprule
            \multirow{2}{*}{Methods} & \multirow{2}{*}{Metric} &\multicolumn{4}{c|}{ETTm2}&\multicolumn{4}{c|}{Electricity}&\multicolumn{4}{c|}{Exchange}&\multicolumn{4}{c|}{Traffic}&\multicolumn{4}{c}{Weather}\\
            & & 96 & 192 & 336 & 720 & 96 & 192 & 336 & 720 & 96 & 192 & 336 & 720 & 96 & 192 & 336 & 720 & 96 & 192 & 336 & 720\\
            \midrule

%%%%%%%%%%%%%%%%%%%%%%%%%%%%%%%%%%%%%%%%% paste start from here~
\multirow{2}{*}{\makecell[c]{TVT \\ (Linear, none)}} &MSE &\textbf{0.181} &\textbf{0.249} &\textbf{0.324} &\textbf{0.485} &\underline{0.157} &\textbf{0.167} &\underline{0.183} &\underline{0.213} &\textbf{0.085} &\underline{0.197} &\underline{0.354} &\textbf{0.846} &\textbf{0.482} &\textbf{0.491} &\textbf{0.505} &\underline{0.543} &\textbf{0.172} &\textbf{0.216} &\textbf{0.263} &\textbf{0.323} \\
 &MAE &\textbf{0.269} &\textbf{0.324} &\textbf{0.381} &\textbf{0.489} &\textbf{0.253} &\textbf{0.265} &\textbf{0.283} &\textbf{0.314} &\textbf{0.212} &\underline{0.340} &\underline{0.460} &\textbf{0.716} &\textbf{0.304} &\textbf{0.309} &\underline{0.319} &\underline{0.337} &\textbf{0.238} &\underline{0.277} &\textbf{0.317} &\textbf{0.360} \\
\midrule
\multirow{2}{*}{\makecell[c]{TVT \\ (Linear, t.e.)}} &MSE &\underline{0.182} &\underline{0.304} &\underline{0.359} &\underline{0.529} &\textbf{0.155} &\underline{0.172} &\textbf{0.182} &\textbf{0.210} &\underline{0.111} &\textbf{0.193} &\textbf{0.348} &\underline{0.859} &\underline{0.489} &\underline{0.495} &\underline{0.506} &\textbf{0.540} &0.267 &0.273 &0.384 &0.417 \\
 &MAE &\underline{0.282} &\underline{0.382} &\underline{0.415} &\underline{0.519} &0.257 &\underline{0.274} &\underline{0.287} &\underline{0.316} &\underline{0.254} &\textbf{0.333} &\textbf{0.459} &\underline{0.726} &\underline{0.310} &\underline{0.314} &\textbf{0.318} &\textbf{0.335} &0.363 &0.352 &0.449 &0.453 \\
\midrule
\multirow{2}{*}{\makecell[c]{TVT \\ (Transformer, none)}} &MSE &0.260 &0.383 &0.490 &0.713 &0.159 &0.187 &0.198 &0.230 &0.320 &0.618 &0.840 &1.347 &0.715 &0.656 &0.670 &0.718 &\underline{0.179} &\underline{0.219} &\underline{0.277} &\underline{0.337} \\
 &MAE &0.358 &0.442 &0.498 &0.592 &\underline{0.255} &0.281 &0.294 &0.327 &0.391 &0.571 &0.685 &0.899 &0.403 &0.369 &0.375 &0.403 &\underline{0.246} &\textbf{0.272} &\underline{0.328} &\underline{0.363} \\
\midrule
\multirow{2}{*}{\makecell[c]{TVT \\ (Transformer, t.e.)}} &MSE &0.218 &0.392 &0.603 &1.311 &0.158 &0.176 &0.201 &0.244 &0.251 &0.543 &0.810 &1.344 &0.732 &0.667 &0.674 &0.758 &0.248 &0.359 &0.541 &0.757 \\
 &MAE &0.326 &0.452 &0.541 &0.892 &0.264 &0.278 &0.304 &0.340 &0.352 &0.547 &0.683 &0.903 &0.427 &0.388 &0.392 &0.418 &0.337 &0.425 &0.546 &0.661 \\
\midrule
\multirow{2}{*}{\makecell[c]{TPT \\ (one forward)}} &MSE &0.339 &0.713 &1.649 &4.034 &0.323 &0.326 &0.347 &0.329 &0.232 &0.746 &1.335 &2.228 &0.786 &0.767 &0.788 &0.776 &0.296 &0.425 &0.632 &0.518 \\
 &MAE &0.410 &0.668 &1.032 &1.564 &0.421 &0.430 &0.448 &0.422 &0.382 &0.694 &0.972 &1.176 &0.462 &0.453 &0.459 &0.460 &0.379 &0.474 &0.595 &0.537 \\
\midrule
\multirow{2}{*}{\makecell[c]{TPT \\ (step by step)}} &MSE &0.311 &0.357 &0.397 &0.571 &0.465 &0.683 &0.990 &1.268 &0.824 &0.900 &0.986 &1.359 &1.236 &1.253 &1.534 &2.042 &0.289 &0.346 &0.423 &0.486 \\
 &MAE &0.408 &0.437 &0.460 &0.598 &0.478 &0.580 &0.710 &0.878 &0.767 &0.802 &0.839 &0.982 &0.644 &0.654 &0.762 &0.960 &0.313 &0.343 &0.381 &0.455 \\
\bottomrule
%%%%%%%%%%%%%%%%%%%%%%%%%%%%%%%%%%%%%%%%%%%%%%%%%%%%%%%%%%%%%%%% paste end here~
            
        \end{tabular}
    }
    \label{tab:basic-cmp}
\end{table*}
\begin{table}[!ht]
    \centering
    \caption{Comparison between TVT and MLP-based models on three datasets with input length $I=96$ and forecasting length $O \in \{96,192,336,720\}$. Lower MSE and MAE indicate better performance. The best results are highlighted in \textbf{bold}.}
    \adjustbox{width=0.9\linewidth}{
        \begin{tabular}{cc|cc|cc|cc}
            \toprule
            \multicolumn{2}{c|}{Methods} & \multicolumn{2}{c|}{TVT (Linear)} & \multicolumn{2}{c|}{MLP} & \multicolumn{2}{c}{MLP-Mixer} \\
            \midrule
            \multicolumn{2}{c|}{Metric} & MSE & MAE & MSE & MAE & MSE & MAE  \\
%%%%%%%%%%%%%%%%%%%%%%%%%%%%%%%%%%%%%%%%%%%%%%%%%%%%%%%%%%%%%%%%%%%%
\midrule
\multirow{4}{*}{\rotatebox{90}{ETTm2}}& 96& \textbf{0.181}& \textbf{0.269}& 0.208& 0.316& 0.200& 0.300\\
~& 192& \textbf{0.249}& \textbf{0.324}& 0.292& 0.375& 0.294& 0.372\\
~& 336& \textbf{0.324}& \textbf{0.381}& 0.393& 0.440& 0.392& 0.435\\
~& 720& \textbf{0.485}& \textbf{0.489}& 0.564& 0.533& 0.573& 0.538\\
\midrule
\multirow{4}{*}{\rotatebox{90}{Electricity}}& 96& 0.157& \textbf{0.253}& 0.181& 0.271& \textbf{0.156}& 0.265\\
~& 192& \textbf{0.167}& \textbf{0.265}& 0.187& 0.279& 0.173& 0.281\\
~& 336& \textbf{0.183}& \textbf{0.283}& 0.199& 0.293& 0.185& 0.294\\
~& 720& \textbf{0.213}& \textbf{0.314}& 0.239& 0.329& 0.250& 0.343\\
\midrule
\multirow{4}{*}{\rotatebox{90}{Traffic}}& 96& \textbf{0.482}& \textbf{0.304}& 0.519& 0.326& 0.531& 0.333\\
~& 192& \textbf{0.491}& \textbf{0.309}& 0.509& 0.327& 0.506& 0.319\\
~& 336& \textbf{0.505}& \textbf{0.319}& 0.525& 0.331& 0.532& 0.327\\
~& 720& \textbf{0.543}& \textbf{0.337}& 0.565& 0.351& 0.572& 0.339\\
\bottomrule
%%%%%%%%%%%%%%%%%%%%%%%%%%%%%%%%%%%%%%%%%%%%%%%%%%%%%%%%%%%%%%%%%%%%
            
        \end{tabular}
    }
    
    \label{tab:mlp-cmp}
\end{table}
\begin{table}[!ht]
    \centering
    \caption{Comparison of inference time (Time, unit is millisecond) and number of parameters (Param., unit is million) among different models on five datasets. Metrics are evaluated as the average of corresponding values under different forecasting length.}
    \adjustbox{width=1.0\linewidth}{
        \begin{tabular}{c|c|ccccc|c}
                \toprule
                Methods & Metric & ETTm2 & Electricity & Exchange & Traffic & Weather & Average \\
                \midrule
%%%%%%%%%%%%%%%%%%%%%%%%%%%%%%%%%%%%%%%%%%%%%%%%%%%%%%%%%%%%%%%%%%%%%%%%%%%%%%%%%%%
\multirow{2}{*}{\makecell[c]{TVT \\ (Linear, none)}} & Time (ms)& 0.862& 0.967& 0.862& 1.000& 0.823 & 0.903\\
~ & Param. (M)& 0.126& 0.126& 0.126& 0.126& 0.126& 0.126 \\
\midrule\multirow{2}{*}{\makecell[c]{TVT \\ (Linear, t.e.)}} & Time (ms)& 0.919& 1.014& 0.908& 1.050& 0.854& 0.949\\
~ & Param. (M)& 0.126& 0.127& 0.126& 0.129& 0.126& 0.127\\
\midrule\multirow{2}{*}{\makecell[c]{TVT \\ (Transformer, none)}} & Time (ms)& 1.739& 1.884& 1.715& 1.938& 1.678& 1.791\\
~ & Param. (M)& 2.174& 2.174& 2.174& 2.174& 2.174& 2.174\\
\midrule\multirow{2}{*}{\makecell[c]{TVT \\ (Transformer, t.e.)}} & Time (ms)& 1.827& 1.977& 1.812& 2.031& 1.773& 1.884\\
~ & Param. (M)& 2.174& 2.177& 2.174& 2.181& 2.174& 2.176\\
\midrule\multirow{2}{*}{\makecell[c]{TPT \\ (one forward)}} & Time (ms)& 2.551& 2.629& 2.541& 2.632& 2.435& 2.558\\
~ & Param. (M)& 5.284& 6.410& 5.288& 8.350& 5.335& 6.133\\
\midrule\multirow{2}{*}{\makecell[c]{TPT \\ (step by step)}} & Time (ms)& 690.5& 708.8& 701.7& 708.9& 667.4& 695.5\\
~ & Param. (M)& 10.52& 10.84& 10.52& 11.40& 10.53& 8.761\\
\bottomrule
%%%%%%%%%%%%%%%%%%%%%%%%%%%%%%%%%%%%%%%%%%%%%%%%%%%%%%%%%%%%%%%%%%%%%%%%%%%%%%%%%%%
            \end{tabular}
    }
    \label{tab:speed-modelsize-cmp}
\end{table}

\begin{figure*}[!ht]
    \centering
    \includegraphics[width=1.\linewidth]{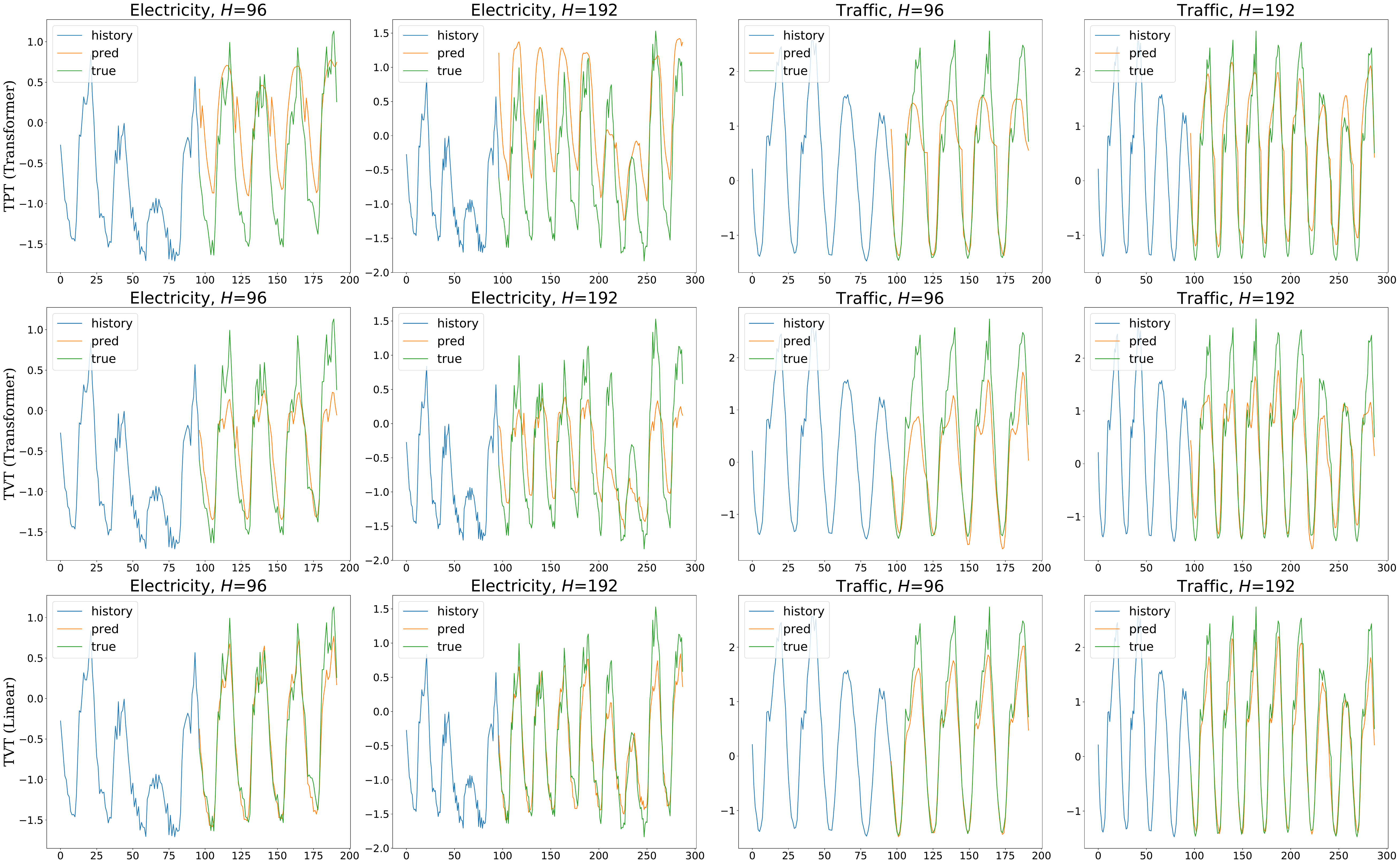}
    \caption{Top to bottom: results generated by a pretrained vanilla TPT Transformer, a pretrained TVT Transformer with the Transformer decoder and a pretrained TVT Transformer with the Linear decoder. Horizon $H=96$ in the first and third columns, $H=192$ in the second and fourth columns. Since the time series are multi-variate, We randomly select one time variable to visualize. It can be observed that the predictions of the vanilla TPT Transformer can barely cling to the ground truth and lacks details like pinnacles and zigzags, which the TVT based Transformers handle better.}
    \label{fig:plot_seqs}
\end{figure*}

\begin{figure*}
    \centering
    \includegraphics[width=1.\linewidth]{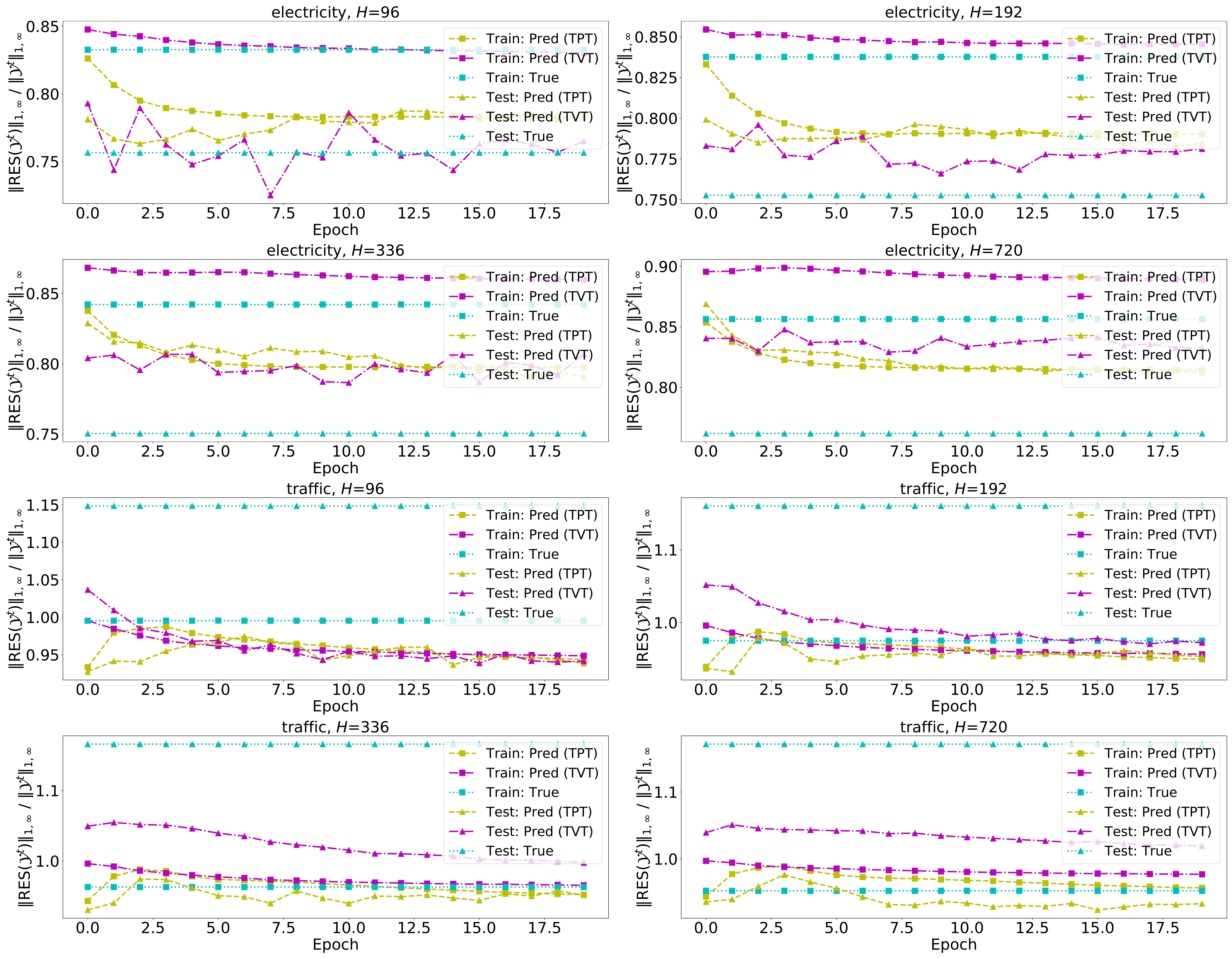}
    \caption{Time point token uniformity $\mathrm{TU}(\mathbf{Y}^t_{TPT})$, $\mathrm{TU}(\hat{\mathbf{Y}}^t_{TVT})$ and $\mathrm{TU}(\mathbf{Y}^t_{gt})$ on both the training set and the test set. We conduct different experiments with horizon $H \in \{96, 192, 336, 720 \}$ and on different datasets (electricity and traffic). A lower ratio value means a higher token uniformity.}
    \label{fig:ratios_ecl-traffic}
\end{figure*}

\section{Experiments}

In this section, we first give our detailed experiment settings. Then, to evaluate the evolved Transformer architectures, we conduct different experiments on five popular benchmarks: \textbf{ETT}~\cite{zhou2021informer}, \textbf{Electricity}, \textbf{Exchange}~\cite{lai2018modeling}, \textbf{Traffic} and \textbf{Weather}. We first compare our TVT based Transformer with existing TPT based state-of-the-art Transformers, which shows that our simple model substantially outperforms these well-designed MTSF Transformers.
Then we validate the predicting performance of TVT based Transformers with different embedding strategies and decoders, together with the conventional step-by-step Transformer and the vanilla MTSF Transformer. We also design two baseline models to verify how well the self-attention mechanism works to connect tokens along the variable dimension.
We also test inference time and number of parameters for each Transformer model.
Then, we explore how well the TVT based Transformer architecture works to reduce the over-smoothing phenomenon found in the TPT based Transformer architecture. We provide plots of predicted time series produced by the vanilla MTSF Transformer and the TVT based Transformers with different decoders. These plots demonstrate that our TVT strategy effectively avoids the over-smoothing phenomenon brought by the vanilla MTSF Transformer.
Finally, we provide a deeper dive into the token uniformity inductive bias, and find that our TVT strategy largely ease the time point token uniformity of predictions.

\subsection{Implementation Details}
Our models are trained using SGD~\cite{sgd} optimizer with learning rates in the range of 5e-1 to 1e-3, which ensures different models in different datasets bring out their best performance. If no loss degradation in the valid set after three epochs, an early stopping will occur. The mean square error (MSE) is used to be the loss criterion. The model dimension $D$ in the TVT based models are simply set to be the observed size $L = 96$. All the deep learning networks are implemented in PyTorch~\cite{paszke2019pytorch} and trained in NVIDIA RTX 3090 GPUs (24GB memory), 4 * Intel Xeon Gold 6240 CPU. The operating system is Ubuntu 20.04.3 LTS.

\subsection{Datasets} 
Here is a detailed description of five experiment datasets: 
(1) \textit{ETT} \cite{zhou2021informer} dataset contains data collected from Electricity Transformers. It contains 7 features, including oil temperature and loads ranging from July 2016 to July 2018.
(2) \textit{Electricity}\footnote{\url{https://archive.ics.uci.edu/ml/datasets/ElectricityLoadDiagrams20112014}} dataset contains the hourly electricity consumption of 321 customers from 2012 to 2014.
(3) \textit{Exchange} \cite{lai2018modeling} records daily exchange rates of eight different countries ranging between 1990 and 2016.
(4) \textit{Traffic}\footnote{\url{http://pems.dot.ca.gov}} is a collection of hourly data from California Department of Transportation, which describes the road occupancy rates measured by different sensors on San Francisco Bay area freeways. 
(5) \textit{Weather}\footnote{\url{https://www.bgc-jena.mpg.de/wetter/}} is recorded every 10 minutes for 2020 whole year, which contains 21 meteorological indicators, such as air temperature, humidity, etc.

\subsection{Comparison with State-of-the-art Transformers}\label{compareSOTA}
For a fair comparison, we follow the experiment settings of FEDformer~\cite{zhou2022fedformer}. The observed window size $L$ is set to be 96, and the prediction lengths for both training and testing are set to be 96, 192, 336, and 720, respectively. The batch size is fixed to be 32.

The experiment results are shown in Table~\ref{tab:formers-cmp}. We use the TVT Transformer with one linear layer decoder and without time stamp group embeddings as our final model. It turns out that compared with those complex and carefully designed Transformers, this simple architecture achieves the lowest MSE scores in 19/20 cases and the lowest MAE scores in 18/20 cases. Compared with FEDformer~\cite{zhou2022fedformer} which has the second best performance, our model achieves a $17.5\%$ relative MSE reduction and $11.7\%$ relative MAE reduction. It is worth noting that except for the ETTm2 dataset, the improvements come to $21.2\%$ and $15.3\%$, respectively. While the Exchange dataset shows little periodicity, our model can still largely exceed the FEDformer.

\subsection{Validation of the Evolved Transformers}
We provide the test results on all the basic Transformer architectures. The experiment results are shown in Table~\ref{tab:basic-cmp}. TVT means the Time Variable Tokenization based Transformer, and TPT means the Time Point Tokenization based Transformer. In the top 4 columns, "Transformer" means the decoder of the TVT Transformer is still a Transformer decoder, and "Linear" means the decoder is replaced by a simple linear layer. "t.e." means using time stamp groups as embeddings, and "none" means there is no embeddings. TPT based Transformers are two types: "one forward" means the vanilla MTSF Transformer that utilizes the one forward operation, and "step by step" means the conventional Transformer that adopts an autoregressive approach. It can be seen that the TVT Transformers (Linear, t.e./none) achieve the best performance in almost all cases. In datasets like electricity and weather, TVT Transformers (Transformer, t.e./none) perform comparable results to TVT Transformer with a linear layer in the short prediction length cases. However, when coming into the long prediction length cases, the performance of TVT (Transformers, t.e./none) quickly meets the degradation, possibly caused by the limited information in the input of the decoder. Due to the over-smoothing phenomena brought by the TPT strategy, the two TPT based Transformers present poor performance in most cases.

\subsection{Validation of the Self-attention Mechanism}
We conduct experiments to demonstrate how well the attention mechanism connects different time variables. As shown in Table~\ref{tab:mlp-cmp}, we design two models as the baseline. "MLP" means the rest MLP layers when removing all the self-attention layers from the TVT Transformer (Linear). It takes no action to connect different time variables. "MLP-Mixer" means an MLP-Mixer~\cite{tolstikhin2021mlp} like model, which is obtained by replacing the self-attention layers in TVT Transformer (Linear) with MLP layers. It connects the different variables with learnable MLP layers. All three models do not employ the time stamp group embedding. It can be seen that TVT Transformer (Linear) easily beats the two baseline models, demonstrating the superior self-attention mechanism's ability to connect different variables and strengthen the correlations among them.

\subsection{Efficiency comparison}
The Comparison of inference time and number of parameters among different basic Transformer architectures are shown in Table~\ref{tab:speed-modelsize-cmp}. Except for the step-by-step TPT Transformer which applies an autoregressive strategy to its decoder, all the other Transformer architectures achieve a small inference time. Particularly, our final model achieves the fastest inference speed with least parameters and substantially exceeds the previous Transformer models.

\subsection{Validation of over-smoothing Reduction}
We visualize the model input, output, and ground truth derived from different datasets, as shown in Fig.~\ref{fig:plot_seqs}. Rows from top to bottom coming from a pre-trained vanilla TPT Transformer, a pre-trained TVT Transformer (Transformer), and a pre-trained TVT Transformer (Linear). We randomly select a one-time variable from the multi-variate time series to visualize. It can be observed that the predicted curves from the vanilla TPT Transformer can barely cling to the ground truth and lacks details like pinnacles and zigzags, which the TVT based Transformers can handle better.

\subsection{A Deeper Dive into the Time Point Token Uniformity}
We present more analysis on the time point token uniformity in the TPT and TVT based Transformers. Inspired by Dong et al.~\cite{rankCollapse2021}, we use the ratio of a prediction's residual to the prediction itself to measure the token uniformity among the predicted time point tokens, which is defined as:
\begin{equation}
    \mathrm{TU}(\mathbf{Y}^t) = \Vert \mathrm{RES}(\mathbf{Y}^t) \Vert_{1, \infty} ~/~ \Vert \mathbf{Y}^t \Vert_{1, \infty}
\end{equation}
A lower ratio value means a higher token uniformity.
We apply this metric to analyze the change of time point token uniformity during the stage of training and inference. We use the TVT based Transformer with a linear layer decoder as a comparison. All the models are trained for 20 epochs without early stopping. Models are trained and tested on the electricity and traffic dataset, which contain the most variable numbers (321 and 862) among all the datasets. In each epoch we calculated the metric on the whole training set and test set, and took the average values as the final results, respectively. Results are shown in Fig.~\ref{fig:ratios_ecl-traffic}.

In Fig.~\ref{fig:ratios_ecl-traffic}, the dotted cyan lines represent the time point token uniformity in ground truth $\mathrm{TU}(\mathbf{Y}^t_{gt})$, the dashed yellow lines represent the time point token uniformity in predictions from a pretrained TPT based Transformer $\mathrm{TU}(\mathbf{Y}^t_{TPT})$, and the dash-dotted magenta lines represent the time point token uniformity in predictions from a pretrained TVT based Transformer $\mathrm{TU}(\hat{\mathbf{Y}}^t_{TVT})$. We can conclude three main points:
\begin{enumerate}
    \item \textbf{The uniformity of time point tokens in the ground truth $\mathbf{Y}^t_{gt}$ varies from the training set to the test set.}
    The uniformity of ground truth time point tokens presents a $10 \sim 20\%$ difference between on the training set and on the test set, which indicates that the uniformity among data points at different times is not stable and varies over time $t$. 
    \item \textbf{With the training epoch increasing, the uniformity of time point tokens in $\mathbf{Y}^t_{TPT}$ becomes stronger and keeps almost the same from the training set to the test set, which is different from the varying token uniformity in the ground truth $\mathbf{Y}^t_{gt}$.}
    With the training epoch increasing, the dashed yellow curves representing different $\mathrm{TU}(\mathbf{Y}^t_{TPT})$ on the training set (represented by the squared points) and the test set (represented by the triangular points) gradually fall and progressively come together. It indicates a consistent time point token uniformity over time $t$, which is different from the varying token uniformity in the ground truth. However, the dash-dotted magenta curves representing different $\mathrm{TU}(\hat{\mathbf{Y}}^t_{TVT})$ on the training set (represented by the squared points) and the test set (represented by the triangular points) are relatively far apart and closer to the ground truth.
    \item \textbf{In the training stage, the uniformity of time point tokens in $\mathbf{Y}^t_{TPT}$ quickly becomes stronger than the uniformity in ground truth $\mathbf{Y}^t_{gt}$, which indicates an over-fitting.} 
    With the training epoch increasing, the dashed yellow curves representing $\mathrm{TU}(\mathbf{Y}^t_{TPT})$ on the training set (represented by the squared points) quickly becomes much lower than the dotted cyan curves representing $\mathrm{TU}(\mathbf{Y}^t_{gt})$, which indicates a high degree of over-fitting. By contrast, in most cases the dash-dotted magenta curves representing $\mathrm{TU}(\hat{\mathbf{Y}}^t_{TVT})$ on the training set fall slower and cling to the ground truth curves closely, showing less over-fitting. Therefore, the curves representing $\mathrm{TU}(\hat{\mathbf{Y}}^t_{TVT})$ on the test set (represented by the triangular points) can get closer to the ground truth. 
\end{enumerate}

It can be seen that the token uniformity inductive bias forces TPT based Transformers' predictions to be consistent along the time dimension both on the training set and the test set, thus resulting in the over-smoothing phenomena in the predictions from TPT based Transformers. By contrast, in most cases the TVT based Transformers can keep the variety of time point tokens along the time dimension, and produce a more accurate prediction.

\section{Conclusion and Future Work}

\paragraph{Conclusion}
This paper explored the basic configurations in existing Transformer architectures designed for multi-variate time series forecasting. We discovered that the current tokenization strategy
in MTSF Transformer architectures ignores the \textit{token uniformity} inductive bias of Transformers. Therefore, the Time Point Tokenization (TPT) based transformers struggle to capture details in time series and present poor performance. 
We change the tokenization strategy into Time Variable Tokenization (TVT) and redesign the decoder structure along with the embedding approach to match the TVT strategy. 
Experiments show that this simple strategy significantly reduces the over-smoothing phenomenon and brings a high-performance improvement.

\paragraph{Limitations and Future Work}
Although the core idea, time variable tokenization strategy, works well in different scenes, the main Transformer architecture is simple to some extent. It is junior to apply MLP layers along the time dimension to connect different time points, and more effective time representation methods like auto-correlation mechanism proposed by \cite{wu2021autoformer} and Frequency enhancing methods proposed by \cite{zhou2022fedformer} can be integrated in to the Transformer architectures. Also, the simple linear layer decoder can be replace by some powerful MLP based MTSF model like N-BEATS~\cite{oreshkin2019n}.

\bibliography{ref}

\end{document}